# Time-Series Forecasting: Unleashing Long-Term Dependencies with Fractionally Differenced Data

Sarit Maitra
Alliance Business School
*Alliance University*
Bengaluru, India
sarit.maitra@gmail.com

Vivek Mishra
School of Applied mathematics
*Alliance University*
Bengaluru, India
vivek.mishra@alliance.edu.in

Srashti Dwivedi
School of Applied Mathematics
*Alliance University*
Bengaluru, India
srashti.dwivedi@alliance.edu.in

Sukanya Kundu
Allince School of Business
*Alliance University*
Bengaluru, India
sukanya.kundu@alliance.edu.in

Goutam Kr. Kundu
Alliance School of Business
*Alliance University*
Bengaluru, India
goutam.kundu@alliance.edu.in

*Abstract*— This study introduces a novel forecasting strategy that leverages the power of fractional differencing (FD) to capture both short- and long-term dependencies in time series data. Unlike traditional integer differencing methods, FD preserves memory in series while stabilizing it for modeling purposes. By applying FD to financial data from the SPY index and incorporating sentiment analysis from news reports, this empirical analysis explores the effectiveness of FD in conjunction with binary classification of target variables. Supervised classification algorithms were employed to validate the performance of FD series. The results demonstrate the superiority of FD over integer differencing, as confirmed by Receiver Operating Characteristic/Area Under the Curve (ROCAUC) and Mathews Correlation Coefficient (MCC) evaluations.

*Keywords— classification; forecasting; fractional difference; mathews correlation; time-series;*

## I. INTRODUCTION

The shocks or fluctuations in the time series can have a lasting impact and affect future values over an extended period. This has garnered significant attention from both academia and practitioners in the field of forecasting. Several researchers, such as Doukhan et al. (2002), Robinson (1995), Mikosch & Stărică (2004), Nguyen et al. (2020), and Zhang et al. (2018), have contributed to this understanding of LRD and its relevance in financial price series forecasting. As we go further back in a time series with short-range dependence, the influence of past values on present values rapidly decreases. However, this impact lasts for a longer time in time series with long-range dependence, frequently leading to a slow decay of autocorrelation functions. This is characterized by the persistence of shocks and the extended influence of past observations. In the past, researchers highlighted the presence of cumulative LRD in time series and claimed that it causes non-linearity (e.g., Kitagawa, 1987; Haubrich 1993; Granger & Joyeux, 1980).

There is a shift towards moving beyond a mere reliance on mean values and non-Gaussian modelling techniques have emerged to represent the underlying patterns and fluctuations in price series. When it comes to price forecasting, researchers like David et al. (2017) and Asl et al. (2017) have emphasized the significance of this modelling approaches in comprehending the complexities of market dynamics. Interestingly, Serinaldi (2010) contends that, despite being noisy, real-world time series display persistent behavior in their observations. This realization has broadened the area of research in this field, with economists recognizing that financial series may exhibit LRD due to stochastic behavior on both current and distant past values. The research landscape has expanded because of the pioneering approaches developed in the past by Granger and Joyeux (1980) and Hosking (1981). In a recent study, Castellano et al., 2020 highlighted the importance of LRD. Their study focused on time series generated by stochastic processes and highlighted the intricate relationship between LRD and the observed autocorrelation patterns. LRD can therefore reveal how firmly systems depend on prior realizations and, subsequently, how quickly they bounce back from positive or negative shocks.

Our argument here centred against integer differencing, which is widely used to stationarize price series. This stationary process resulted in important series memory loss, which is critical for the predictive power of a model. Ayadi et al. (2009) proposed fractional integration as a solution to the complexity of modelling time series. Financial time series often exhibit long-range dependence, which means that past values have a significant impact on future values over extended time horizons. This property can make modeling and forecasting challenging. FD can be used to reduce the long-term dependencies in the data by applying a fractional difference that is less than 1. This can make the data more amenable to modeling with traditional methods that assume shorter memory or independence. Through this work, we aim to determine the best difference strategy that preserves crucial series memory while enabling successful prediction of future observations by studying the trade-off between stationarity and memory.

In the past, Hosking (1981) introduced an autoregressive moving average model with FD to conceptualize the idea of LRD in time series. In recent times, Mills (2019) reiterated the same concept by investigating medium- to long-term forecasting. They concluded that being able to spot recurring patterns in a time series would be quite helpful. The relationships between a current value $x_t$ of a series and a set of lagged values $x_{t-k}$, where $k = 1, 2, or\ 3$, are commonly referred to as autocorrelations. The definition of the lag-k autocorrelation is:

$$r_k = \frac{\sum_{t=k+1}^{t}(x_t - \bar{x})(x_{t-k} - \bar{x})}{T s^2} \qquad (1)$$

where the sample mean and variance of $x_t$ are denoted by, $\bar{x} = T^{-1}\sum_{t=1}^{T} x_t$ and $s^2 = T^{-1}\sum_{t=1}^{T}(x_t - \bar{x})^2$, respectively. The autocorrelation function (ACF), a collection of sample correlations for various values of k, is essential for time-series research.

Through this work, we present an empirical analysis and make a theoretical contribution by incorporating theoretical aspects into the experimentation. We re-examine the relationship between stationarity and memory using their concept to show that raising the level of differentiation leads to stationarity, but at the expense of fundamental memory loss. While prior researchers have recorded a few great mathematical theories concerning FD time series, we are aware of no published studies that involve forecasting using real-world price series and binary classifications in the applied field. We have used long range dependence and long memory interchangeably in this work.

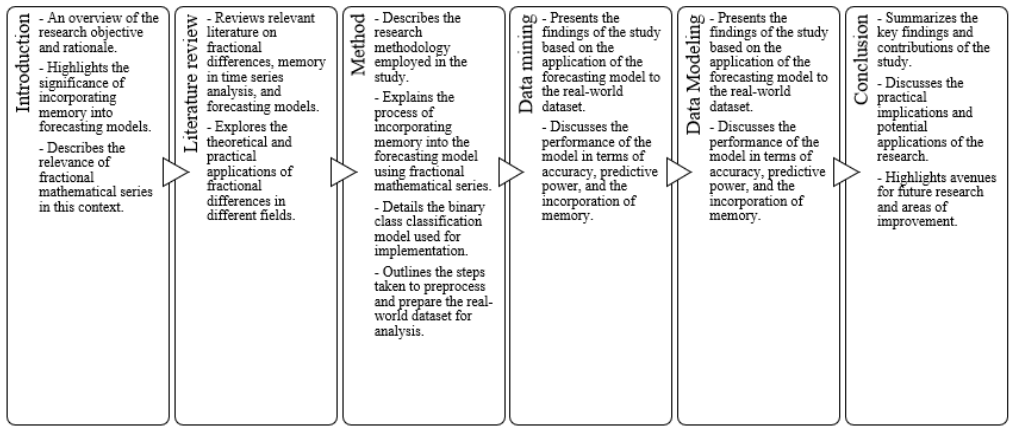

Fig. 1. Workflow diagram

Fig. 1 presents a flow diagram of the body of this study. It explains the different sections covered in this study and provides a summary of each section.

## II. LITERATURE REVIEW

Fractional differentiation (FD) and long-memory processes have indeed been subjects of considerable research in the fields of econometrics and time-series analysis (for example, Granger & Joyeux, 1980; Hosking, 1981; Beran, 1994; Baillie, 1996; Granger & Ding, 1995; Diebold & Inoue, 2001). Hurst (1951, 1957), Mandelbrot & Wallis (1968), McLeod & Hipel (1978), and Smith & Harris (1987) were among the first to study LRD using hydrogeological data. Despite their contributions to the corpus of knowledge, most of these studies have concentrated on theoretical concepts and mathematical formulations without addressing how they might be implemented in practical circumstances. Granger & Joyeux (1980) and Hosking (1981) were pioneers in linking long memory to FD. They presented fractional finite difference and Grunwald-Letnikov FD, stating that, if $Y_t$ = stochastic process, $b$ = lag operator such that $b(Y_t) = Y_{t-1}$, $d = FD$, and $\varepsilon_t$ = white noise. Baillie (1996) also argued the possibility of performing time-series modelling, considering that parameter $d$ assumes non-integer values.

The theoretical aspects of FD were developed by Johansen and Nielsen (2014, 2019 and 2010), who offered mathematical formulations and investigated the features of those formulations. Gil-Alana et al. (2017) and MacKinnon (1996) offered more information on the statistical characteristics and estimating techniques connected to FD. In the area of econometrics, Kapetanios et al. (2019) and Cavaliere et al. (2017, 2022)'s work assisted us in understanding how FD might be used to examine time-series data in the economy and finance. In an earlier study, Sadaei et al. (2016) provided an empirical analysis highlighting the benefits and potential of FD in a financial price series. Models like the Fractionally Integrated GARCH (FIGARCH) model are used to capture long-memory behavior in financial data (Chen et al., 2022; Pan et al., 2023).

Flores-Muoz et al. (2018) and David et al. (2017) used autoregressive models to incorporate FD into price and commodity series. Wang and Xu's (2022) work also emphasized FD are preferred than integer differences in equations which strengthens the case for FD. The relationship between non-stationary time series and their stationary transformations was explored by De Prado (2018) to justify the occurrence of memory loss. This investigation specifically involved comparing first-order logarithmic series and FD. All these works have collectively contributed to the development of FD theory and its applications in our work.

## III. METHODOLOGIES

Assuming that $Y_t$ is the result of taking the $d^{th}$ difference of a time-series $X_t$, where $t = 0, 1 \ldots, (n-1)$. With that assumption, it is possible to explain the backward difference operator as:

$$Y_t = \Delta^d X_t = (1-b)^d X_t, \text{ where } \Delta = (1-b) \quad \ldots (2)$$

Here, $\Delta$ (Backward difference operator) represents differencing, $\Delta^d X_t = (1-b)^d X_t$ expresses the backward difference operator, where b represents the lag operator. The lag operator shifts the series backward by one step ($bX_t = X_{(t-1)}$). Our argument here is that, instead of just subtracting the previous observation, we subtract a weighted combination of past observations. These weights are determined by the FD parameter $(d)$, which controls the degree of persistence or memory in the series.

Hosking, (1981) described FD as the discrete-time equivalent of stochastic movement, using the backward shift operator for this purpose. In the case of $d \in (0, 1)$, the time-series ($Y_t$) shows long-memory.

In the case of a 1st order difference *(d = 1)*,

$$Y_t = (1-b)X_t = X_{t-b} X_t = X_t - X_{t-1} \quad (3)$$

Likewise, in the case of $d = 2$, the 2nd degree polynomial can be calculated as:

$$Yt = (1 - b)2Xt$$

$$= (1 - b)(X_t - X_{(t-1)}) = X_t - 2X_{(t-1)} + X_{(t-2)} \quad (4)$$

The $d^{th}$ difference for any integer d can be defined by extending $(1 - b)^d$ and then applying the resulting polynomial in b to $X_t$.

The coefficients (weights) in the FD formula can be derived using Taylor series expansion and the gamma function. These coefficients determine the contribution of each lagged observation to the current value of the differenced series.

$$(1-b)^d = 1 + \frac{d}{1!}(-b)^1 + \frac{d(d-1)}{2!}(-b)^2 + \frac{d(d-1)(d-2)}{3!} \quad (5)$$

$$= \sum_{j=0}^{\infty} \frac{d(d-1)(d-2)\ldots(d-(j-1))}{j!}(-1)^j b^j \quad (6)$$

The numerator in the above expression has $j$ factors, except when $j = 0$, the sign of each factor in the numerator is now changed by multiplying it by $-1$:

$$(1-b)^d = \sum_{j=0}^{\infty} \frac{-d(1-d)(2-d)\ldots((j-1)-d)}{j!} b^j \quad (7)$$

After that, multiply by $1 = \frac{\Gamma(j-j-d)}{\Gamma(-d)}$ and by switching the elements' positions, the following equation was formulated:

$$(1-b)^d = \sum_{j=0}^{\infty} \frac{(j-1-d)(j-2-d)\ldots(j-j-d)\Gamma(j-j-d)}{j!\,\Gamma(-d)} b^j \quad (8)$$

The recurrence property of the gamma function is then used to: $\Gamma(X) = (X-1)\Gamma(X-1)$, the numerator can be expressed as $\Gamma(j-d)$. Thus, the formula can be revised as commonly used representation of the FD operator: $(1-b)^d = \sum_{j=0}^{\infty} \frac{\Gamma(j-d)}{\Gamma(j+1)\Gamma(-d)} b^j$.

It is important to calculate the coefficients in the series to perform a FD algorithm: $\omega_j = \frac{\Gamma(j-d)}{\Gamma(j+1)\Gamma(-d)}$, $j = 0, 1, 2 \ldots$ Because these coefficients are used to multiply observations in the time-series, this unending line of coefficients may be condensed to the length of the data series. When calculating these coefficients, there is a challenge since the numerator and denominator grow to enormous sizes and the computer cannot handle them. The recursive feature of the gamma function was used to generate a recursive formula for the $\omega_j$:

$$\omega_0 = \frac{\Gamma(0-d)}{\Gamma(1)\Gamma(-d)} = 1 \quad (9)$$

$$\omega_j = \frac{\Gamma(j-d)}{\Gamma(j+1)\Gamma(-d)} = \frac{(j-d-1)\Gamma(j-d-1)}{j\Gamma(j)\Gamma(-d)} = \frac{(j-d-1)}{j} \omega_{(j-1)} \quad (10)$$

In the recursive formula for $\omega_j$, the gamma function is not used; hence, it is possible to calculate $\omega_j$ for extremely large values of j. The only computation needed is to multiply $\omega_{(j-1)}$ by $(j-d-1)/j$. For the series to keep its memory with a real non-integer positive d, we have $Y_t$ = cumulative sum with weights $\omega_j$ and values X is formulated as:

$$Y_t = \sum_{j=0}^{\infty} \omega_j X(t-j) \quad (11)$$

where,

$$\omega_j = \left\{1, -d, \frac{d(d-1)}{2!}, \frac{d(d-1)(d-2)}{3!}, \ldots, (-1)^j \prod_{i=0}^{j-1} \frac{d-i}{j!}, \ldots\right\} \quad (12)$$

and

$$X = \{X_{(t-k)}, \ldots\} \quad (13)$$

The weights $\omega_j$ determine the contribution of each lagged observation $X_{(t-j)}$ to the current value $Y_t$. However, the series is theoretically infinite, which means it includes an infinite number of terms. In practice, it is not feasible to include all infinite terms when calculating $Y_t$. Therefore, a threshold is introduced to truncate the series and include only a finite number of terms.

## IV. DATA MINING

The CRISP (CRoss Industry Standard Process) data mining procedure is followed here, except for the last phase (deployment). The SPY series is considered here, which is an exchange-traded fund (ETF) that tracks the performance of the SP500 index. Given the substantial volume of SPY, its changes can present a stock market trend. Daily data from 1 January 2010 till 6 November 2020, i.e., 2627 datapoints, were taken with initial regular parameters, e.g., Open, High, Low, Adj. Close, Volume. Table 1 displays the statistical summary of the dataset.

TABLE I. STATISTICAL SUMMARY OF DATASET

|  | Open | High | Low | Adj Close | Volume |
|---|---|---|---|---|---|
| count | 2627 | 2627 | 2627 | 2627 | 2627 |
| mean | 200.50 | 201.52 | 199.41 | 175.76 | 127173300 |
| std | 61.27 | 61.53 | 61.00 | 61.27 | 75498200 |
| min | 103.11 | 103.41 | 101.12 | 79.87 | 20270000 |
| 25% | 140.45 | 140.81 | 139.69 | 114.02 | 75175350 |
| 50% | 201.50 | 202.52 | 200.16 | 173.91 | 107068100 |
| 75% | 250.20 | 254.05 | 249.38 | 231.01 | 157538200 |
| max | 337.79 | 339.07 | 337.48 | 320.12 | 717828700 |

Fig. 2 displays the volatility of the SPY Volume series over the last 252 datapoints. The SP500 index consists of Fortune 500 companies, and the top 35 businesses account for 48% of the index's value. The average emotion generated by the news report portrays the mood of the SPY. Researchers (e.g., Sun et al., 2016, Yang et al., 2018, Bozanta et al., 2021, Obaid & Pukthuanthong, 2022, Chang et al., 2021) have proved the fundamental connection between investor sentiment and stock trends, e.g., bullish, or bearish.

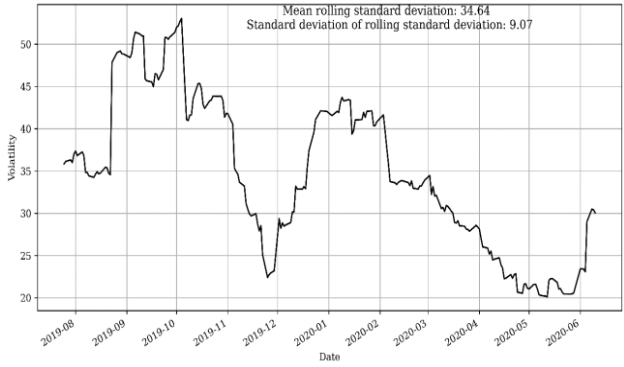

Fig. 2. SPY volatile series

Consequently, sentiment analysis was performed on the daily news headlines about 35 companies throughout a ten-year period, from 2010 to 2020. We derive the daily, continuously compounded rate of return on SPY for the index labelled i as $r_i(t) = ln\left(\frac{x_i(t)}{x_i(t-1)}\right)$, where $x_i(t)$ is the close price of the day t and $x_i(t-1)$ is the close price of the day $t-1$. The compounded return value ($r_i t$), over time is depicted in Fig. 3. The spectrum created by the linear transformation is almost consistent throughout a wide frequency range.

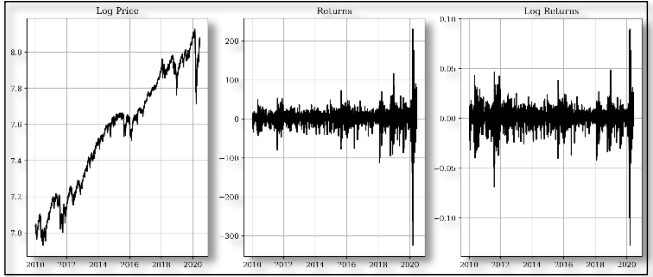

Fig. 3. Temporal evolution of the daily continuously compounded rate of return, $r_i(t)$,

This is analogous to a stochastic, stationary signal, such as white noise. Because there is no linkage with earlier data, each new data value gives the same amount of added information. Because these signals are not the best for exposing dynamical relationships, the daily original price $x_i(t)$ was chosen. The line plot on the extreme right (Fig. 3) displays the time evolution of the daily closing price. The noisy and chaotic-like characteristics are shown in Fig. 4.

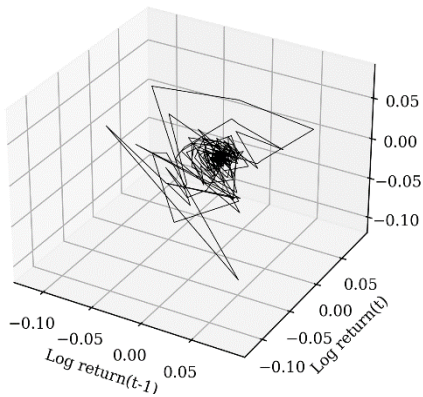

Fig 4. Daily return chaos plot (z level = log returns(t+1))

1) *Hurst Exponent:* The Hurst exponent (HE) used to assess the persistence of time series. Table II displays the HE values at different lags.

TABLE II. HURST EXPONENT AT DIFFERENMT LAG VALUES

| HE with 5 lags | 0.6041 |
|---|---|
| HE with 10 lags | 0.5624 |
| HE with 20 lags | 0.6047 |
| HE with 100 lags | 0.3634 |

HE with 5 lags (0.6041) suggests a moderate level of LRD in the data. HE with 10 lags (0.5624) indicates a slightly weaker long-term dependence. The influence of past values may extend over a slightly longer time scale at this stage. HE with 20 lags (0.6047) is like the 5 lags case, suggesting a consistent level of LRD. Past values continue to influence future values over a moderate time scale. HE with 100 lags (0.3634) indicates a decrease in LRD. The results suggest that the series may have a combination of both short-term and long-term dependencies. The persistence observed in the HE values implies that past values of the series can provide some predictive power or influence future values.

*B. Iterative estimation*

For positive j and $\omega_0 = 1$, weights can be generated iteratively as $\omega_j = -\omega_{j-1}\frac{d-j+1}{j}$. Fig. 5 displays the changes in weight ($\omega$) with different fractional orders: d ∈ (0, 1), $\omega$ = weight for each data sample was estimated for each day of SPY index price and plotted for comparison.

In the case of $d = 0$, all weights are 0 except for $\omega_0 = 1$ where the differenced series overlaps with the original series, and in the case of d = 1, the weights are 0 except for $\omega_0 = 1$ and $\omega_1 = -1$ which show first-order integer differentiation. Since d ∈ [0, 1], all the $\omega$ after $\omega_0$ ($\omega_0 = 1$) are negative and stronger than −1. $\omega$ was decided here by the level of FD to be done.

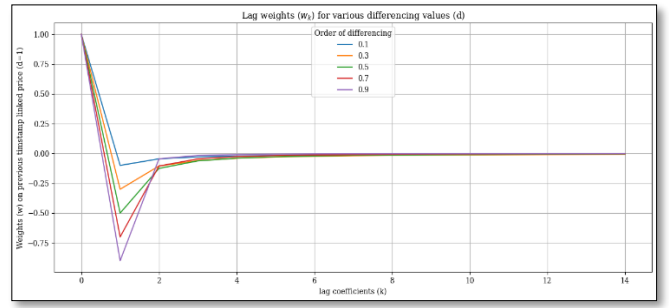

Fig. 5. Lag weights ($w_j$) for various differencing values (d).

The absolute value of $\omega$ can be represented as:

if $\omega_{(j-1)} \neq 0$:

$$\left|\frac{\omega_J}{\omega_{J-1}}\right| = \left|\frac{(d-j+1)}{J}\right| < 1$$

else: $\omega_j = 0$       (14)

When d is +ve and j < (d+1), we obtain that $\frac{(d-j+1)}{j} \geq 0$; this changes the initial $\omega$ sign to alternate. When, $d \in [0, 1]$, int(d) = 0, all weights are negative following $j \geq 1$ {$j \geq d+1$}. It can be inferred that, $\lim_{J \to \infty} \omega_J = 0^-$ when int[d] is even, and $\lim_{J \to \infty} \omega_J = 0^+$ when int[d] is odd. It follows that in the case $d \in (0, 1)$, this means that $-1 < \omega < 0, \forall j > 0$. This change in $\omega$ was needed to achieve $\{Y_t\}_{t=1,\cdots T}$ stationary because memory deteriorates with time.

To simplify the above, in the case of integer differencing orders, such as d = 1, the coefficient of the first lag (lag 1) is exactly -1 because we are subtracting the previous observation. The coefficients for the remaining lags are zero because they are not involved in the differencing process. However, when the differencing order is fractional (e.g., d = 0.5), the coefficients of the lags are no longer exactly -1 for lag 1 and zero for the remaining lags. Instead, each lag has a weight, and these weights converge to zero. Higher orders of differencing typically lead to faster convergence towards zero.

Fig. 6 displays a comparison plot. The plot allows for visual comparison of the effects of differencing on both the original and logarithmic series.

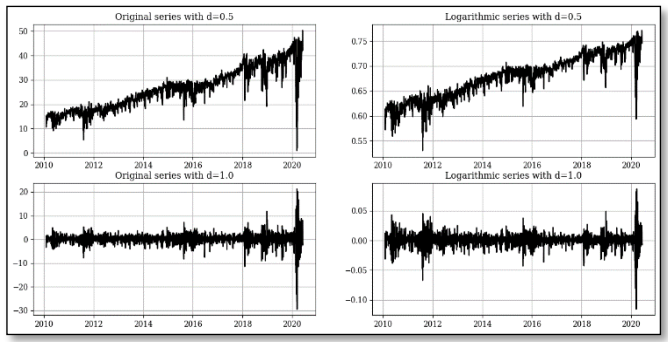

Fig. 6. Comparison of FD on Original and Logarithmic Series

We see that, the higher the differencing order, the more stationary the series becomes, indicating the removal of long-term dependencies and trends. The logarithmic transformation provides additional stability to the series and reduces the influence of extreme values. The p-values for the ADF (Augmented Dicky Fuller) test were calculated for $d \in (0,1)$ by setting a low threshold ($\rho = $ 1e-4) and plotted as shown in Fig 7.

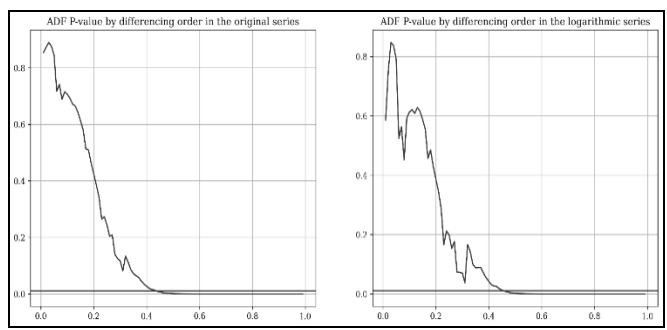

Fig. 7. ADF p-values for traditional differenced series

Fig. 7 depicts both the original and logarithmic series break $\rho$ for the ADF p-value of about 0.4, showing that an integer is not necessary to achieve stationarity. FD acts as a filter to make the series stationary and keep the maximum possible mathematical memory.

To have more clarity on $\rho$ vs. difference values, impact analysis was conducted with 130 combinations of $\rho$ values $\{1e-3, 9e-4, 7e-4, 5e-4, 3e-4, 1e-4, 9e-5, 7e-5, 5e-5, 3e-5\}$ and difference values $\{0.8, 0.75, 0.7, 0.65, 0.6, 0.55, 0.5, 0.45, 0.4, 0.35, 0.3, 0.25, 0.2\}$. Fig. 8 displays the heatmap plot showing the parameter change of the differenced series with various fractional values.

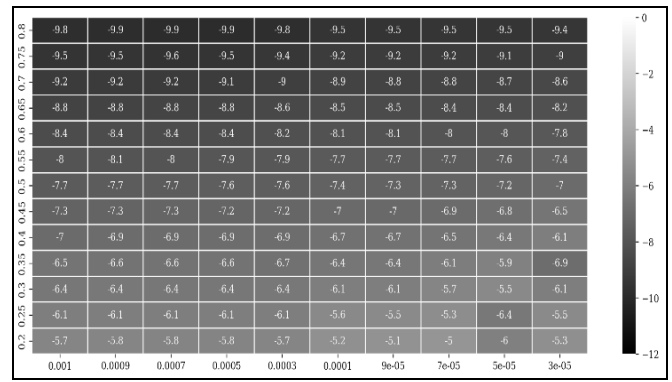

Fig. 8. Heatmap of 130 combinations of threshold values and diff values with ADF test-statistics

It can be concluded that trend-stationarity can be reached without significantly altering a series. If $\rho$ is increased, the test statistics improve marginally because the FD series has access to more data points. When all characteristics are transferred via fractional derivative, certain data points will be reduced. The values of d and threshold can be used to modify the number of remaining data points. We have employed a threshold value of 1e-4.

*C. Optimal d value*

The weights were applied to each data value based on relative weight loss, to be $\lambda_l = \frac{\Sigma_{j=T-l}^{T} |\omega_j|}{\Sigma_{i=0}^{T-1} |\omega_i|}$ as the memory at every point will be different as per the data availability; so, with $d \in (0, 1)$ the amount of memory to be preserved can be decided. Here, the number of historical data points were taken as a fixed window; $\rho$ was configured to fix its length, and data points outside this window were all removed. The ideal value of d was found from the ADF test statistic in Table III.

TABLE III. ADF TEST ON FD VALUES (0-1)

| Diff order | ADF stat | p-value | correlation |
|---|---|---|---|
| 0.0 | -0.358624 | 0.9173 | 0.985639 |
| 0.1 | -1.223851 | 0.6824 | 0.988914 |
| 0.2 | -2.441901 | 0.1410 | 0.974283 |
| 0.3 | -4.396641 | 0.0002 | 0.933779 |
| 0.4 | -7.386350 | 0.0000 | 0.861904 |
| 0.5 | -11.571626 | 0.0000 | 0.742919 |
| 0.6 | -16.612970 | 0.0000 | 0.586000 |
| 0.7 | -22.310789 | 0.0000 | 0.420369 |
| 0.8 | -27.780727 | 0.0000 | 0.275372 |
| 0.9 | -33.023656 | 0.0000 | 0.153951 |
| 1.0 | -37.714796 | 0.0000 | 0.021846 |

Table III shows that, at $d = 0.3$, the FD series passes the ADF test, $pVal = 0.0002 < 0.05$, which is quite early in the differentiation process with a correlation of 93%. ADF test statistics and the (linear) correlation to the original series with different orders of differencing have been displayed to show the trade-off between stationarity and memory. Fig. 9 displays the shape of the original and FD series plotted along the original price and differenced price axes. The low p-value (0.000, Table IV) proves that the data has neither a unit root nor a non-stationary trend. KPSS (Kwiatkowski Phillips Schmidt Shin) p-value < 0.05 rejects $H_0$ around a level, showing the white noise around a trend and showing a stationary process.

TABLE IV. ADF TEST STATISTIC FOR FRACTIONALLY DIFFERENCED SERIES (D = 0.3)

| | Test statistic | p-value | Critical value (1%) | Critical value (5%) | Critical value (10%) |
|---|---|---|---|---|---|
| Adf test statistics | -5.055 | 0.000 | -3.436 | -2.864 | -2.568 |
| KPSS test statistic | 7.398 | 0.010 | 0.739 | 0.463 | 0.347 |

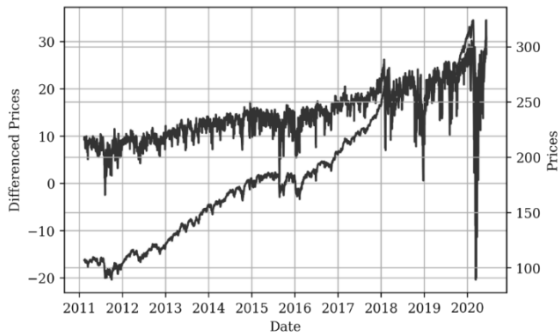

Fig. 9. Shape of original (logarithmic) & FD (0.3) series

The series at this stage displays statistical characteristics that are independent of the time point and preserves much more memory. LRD has proven to have high persistence in data from an empirical standpoint. This satisfies findings from de Prado (2018) that all price series achieve stationarity at around $d < 0.6$, and most of them are stationary even at $d < 0.3$.

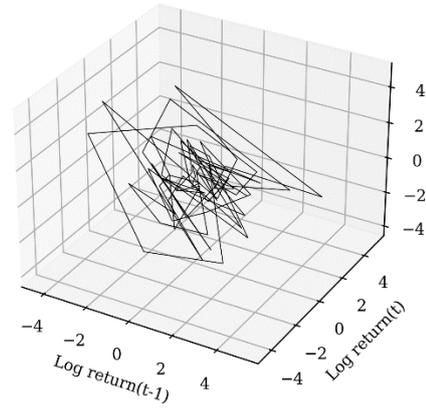

Fig. 10. Chaotic attractor plot (z label=FD with order 3)

Fig. 10 displays 3D chaos plot with FD (0.3). Here, the points in the plot appear to be more clustered, closer together, and follow some discernible patterns; it suggests less chaos compared to Fig. 4 with more scattered and random-looking points.

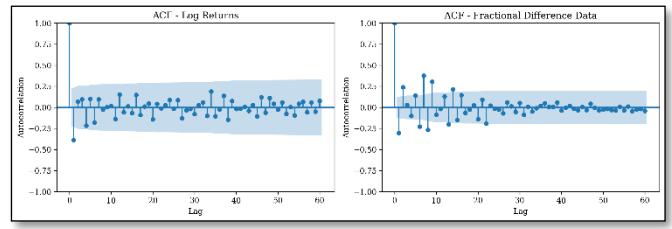

Fig. 11. Autocorrelation plots

Fig. 11 provides an insight into the temporal structure and autocorrelation patterns of the data. There was no significant dependence between the current and previous observations at different time lags. This suggests that log returns exhibit random behavior in the short term, without any noticeable patterns. The FD series displays the persistence of autocorrelation at longer lags. This indicates short- and long-range dependencies in the data. The autocorrelation patterns suggest that past values of the FD data can provide information and influence future values, indicating the presence of some underlying structure or trends in the data.

Fig. 12 displays the boxplots of OHLC series. Asymmetry in the distribution implies that the data is skewed, meaning it is not symmetrically distributed around the mean. This skewness introduces nonlinear transformations to the series. Excessive shortness in the distribution indicates the presence of heavy tails or outliers in the data. These observations suggest that FD has influenced the statistical properties of the historical series.

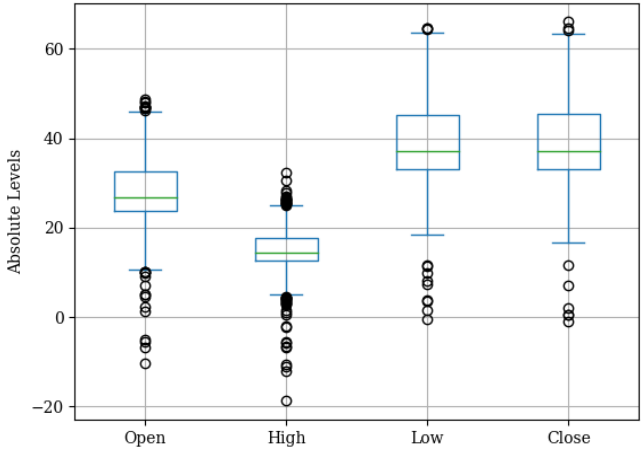

Fig. 12. Boxplots of FD series (OHLC)

## V. MODELLING

The SPY sentiment was computed as the average sentiment score of 35 companies for each day.

$$Sentiment_{SPY}^{(t)} = \frac{1}{N}\sum_{i=1}^{N} Sentiment_{(i)}^{(t)} \quad (15)$$

Data, including date, price movement, closing price, traded volume, and sentiment score, were used to predict future price directions. Depending on the sign of the difference, the volume movement was chosen either as -1 or as 1. The goal is to accurately forecast whether the volume will increase or decrease daily. Daily price changes were included as an additional predictor in the dataset.

$$\text{Daily Price Change} = \frac{ClosePrice(t) - OpenPrice(t)}{OpenPrice(t)} \quad (16)$$

The target variable, which is the outcome of the future direction of Volume was computed as Outcome, Next Day Direction = Volume(t) (backward difference) – Volume(t), where Outcome, Next Day Direction > 0 or = 1 else -1 which makes the target a binary class target.

Volume is the number of shares or contracts traded and is crucial for analyzing market dynamics. Spikes in volume indicate increased activity and interest. By examining the relationship between buying and selling volumes, resistance and support levels can be identified. Resistance occurs when selling volume surpasses buying volume, indicating higher supply than demand. Support happens when buying volume exceeds selling volume, signaling more demand than supply. Understanding volume helps gauge market sentiment and identify potential price levels where assets struggle or find support.

The sentiment score training data is shown in a scatter plot depicted in Fig. 13. A data point's positive score is represented on the x-axis and its negative score is represented on the y-axis. The neutral result is represented by the size of the ball. Particularly near the Centre of the graph, there appears to be a fair amount of randomness. The areas where one of the two sentiment groups predominates over the other can be clearly seen in Fig. 11.

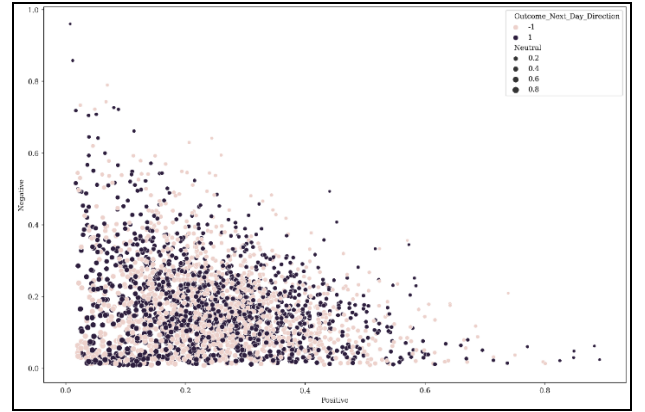

Fig. 13. Data distribution in space

Table V displays the accuracy metrics of different machine learning (ML) models. One of the most important activities in statistics and ML is binary classification, although there is still no broad agreement among scientists on the statistical indicator for assessing binary-class confusion matrices with true positives (TP), true negatives (TN), false positives (FP), and false negatives (FN).

Metrics used here are:

- Accuracy = Predicted output / Actual observation ≈ (TP + TN) / (TP+TN+FP+FN)

- Recall = TP / (TP+FN)

- Precision = TP / (TP + FP)

- ROCAUC: can be viewed using a ROC (Receiver Operating Characteristic) curve showing the variation at each point between TP rate and FP rate. ROCAUC is important here because equal weight was given to both classes' prediction abilities.

- MCC = $\frac{TP*TN - FP*FN}{\sqrt{(TP+FP)*(TP+FN)*(TN+FP)*(TN+FN)}}$

where, (worst value = −1; best value = +1), $Cov(c,l)$ is the covariance of the actual classes c and predicted labels l, and $\sigma_c$ and $\sigma_l$ are the standard deviations, respectively.

- Cohen's Kappa = $\frac{\text{Observed value} - \text{Predicted value}}{1 - \text{Predicted value}}$, where recommended values are = slight: 0 to 0.2, fair: > 0.2, moderate: > 0.4 – 0.6, substantial: > 0.6 – 0.8 and perfect: < 0.80 (Landis & Koch, 1977).

Matthew's correlation coefficient (MCC) is a more dependable statistical measure that only yields a high score if the prediction performed well in each of the four categories of the confusion matrix, proportionally to the size of the dataset's positive and negative elements (Chicco et al., 2021).

KNN and RF can handle non-linear relationships in the data more effectively than others like LogReg and SVM. KNN and RF have the ability to capture complex patterns and interactions, which could be advantageous for modelling non-linearity.

TABLE V. CLASSIFICATION ACCURACY OF DIFFERENT ALGORITHMS WITH INTEGER DIFFERENCE

|  | ML model | Accuracy | ROCAUC | Precision | Recall | MCC | Kappa |
|---|---|---|---|---|---|---|---|
| 1 | LogReg (solver = liblinear) | 62.20% | 64.98% | 60.67% | 62.25% | 30.04% | 29.94% |
| 2 | k-nearest neighbour (k = 2) | 53.62% | 56.45% | 55.85% | 53.33% | 14.43% | 12.96% |
| 3 | k-nearest neighbour (k = 5) | 55.97% | 59.19% | 55.54% | 55.96% | 18.38% | 18.38% |
| 4 | k-nearest neighbour (k = 10) | 57.50% | 58.13% | 58.77% | 57.39% | 16.57% | 16.29% |
| 5 | k-nearest neighbour (k = 50) | 60.81% | 62.05% | 60.76% | 60.78% | 24.11% | 24.11% |
| 6 | k-nearest neighbour (k = 100) | 61.67% | 63.03% | 60.81% | 61.70% | 26.06% | 26.06% |
| 7 | k-nearest neighbour (k = 200) | 62.35% | 63.05% | 60.94% | 62.39% | 26.12% | 26.09% |
| 8 | SVM (kernel = linear) | 62.54% | 65.98% | 60.03% | 62.65% | 32.25% | 31.91% |
| 9 | SVM (kernel = sigmoid) | 53.62% | 57.02% | 53.08% | 53.62% | 14.16% | 14.16% |
| 10 | SVM (kernel = rbf) | 61.72% | 65.75% | 60.24% | 61.78% | 31.60% | 31.48% |
| 11 | SVM (kernel = poly) | 61.15% | 62.35% | 61.06% | 61.13% | 25.11% | 24.75% |
| 12 | RForest (entropy, nodes = 2) | 62.71% | 63.79% | 61.10% | 62.77% | 27.58% | 27.58% |
| 13 | RForest (entropy, nodes = 3) | 62.86% | 63.60% | 61.99% | 62.88% | 27.19% | 27.19% |
| 14 | RForest (entropy, nodes = 5) | 63.13% | 63.59% | 62.34% | 63.15% | 27.19% | 27.19% |
| 15 | RForest (entropy, nodes = 10) | 63.56% | 64.54% | 62.95% | 63.56% | 29.10% | 29.09% |
| 16 | RForest (entropy, nodes = 15) | 62.75% | 64.16% | 63.64% | 62.70% | 28.33% | 28.33% |
| 17 | RForest (entropy, nodes = 20) | 63.86% | 64.90% | 63.69% | 63.85% | 29.90% | 29.82% |
| 18 | RForest (entropy, nodes = 50) | 63.67% | 66.07% | 63.65% | 63.65% | 32.17% | 32.15% |
| 19 | RForest (gini, nodes = 2) | 61.98% | 63.99% | 59.76% | 62.08% | 27.98% | 27.97% |
| 20 | RForest (gini, nodes = 3) | 63.29% | 63.61% | 61.73% | 63.34% | 27.23% | 27.22% |
| 21 | RForest (gini, nodes = 5) | 63.44% | 63.41% | 62.19% | 63.48% | 26.82% | 26.82% |
| 22 | RForest (gini, nodes = 10) | 63.52% | 64.17% | 63.23% | 63.51% | 28.35% | 28.35% |
| 23 | RForest (gini, nodes = 15) | 63.75% | 65.10% | 63.35% | 63.74% | 30.26% | 30.22% |
| 24 | RForest (gini, nodes = 20) | 62.75% | 64.73% | 62.24% | 62.75% | 29.48% | 29.47% |
| 25 | RForest (gini, nodes = 50) | 63.48% | 67.02% | 63.56% | 63.46% | 34.09% | 34.06% |

*LogReg: Logistic regression, **SVM: Support Vector Machine, ***RForest: Random Forest

For the FD series, a noticeable overall improvement was observed. Table VI shows the top scores for each category. MCC was considered here over all the other scores as it is a more balanced approach to classifier assessment, no matter which class is positive or negative.

The KNN model with k = 200 has a relatively high accuracy of 75.85% and a comparable ROCAUC of 73.84%. KNN algorithm is a non-parametric, instance-based learning method that does not assume any specific functional form for the underlying relationship between features and the target variable.

TABLE VI. CLASSIFICATION ACCURACY OF THE FRACTIONAL DIFFERENCED SERIES

|  | ML model | Accuracy | ROCAUC | Precision | Recall | MCC | Kappa |
|---|---|---|---|---|---|---|---|
| 1 | LogReg (solver = liblinear) | 75.90% | 73.84% | 74.98% | 75.86% | 47.69% | 47.67% |
| 2 | k-nearest neighbour (k = 200) | 75.85% | 73.84% | 76.70% | 75.86% | 47.69% | 47.67% |
| 3 | SVM (kernel = poly) | 76.22% | 71.96% | 75.53% | 87.35% | 44.00% | 43.88% |
| 4 | RF (criterion = gini, nodes = 15) | 75.27% | 75.22% | 78.05% | 75.33% | 49.53% | 49.40% |

The precision, recall, MCC, and Kappa values are consistent across both the LogReg and KNN models indicate that both models have similar performance in terms of correctly identifying positive instances (precision), correctly classifying true positive instances (recall), and overall agreement with the true labels (MCC and Kappa).

This study adds to the existing literature by using a formal model setup based on the FD series. In an environment with fractionally differenced integrated variables (d), this method may serve as a statistical framework for examining and differentiating between short- and long-memory effects. Even when prices are intended to follow a random walk, sentiment scores can improve the accuracy with which statistical models predict stock-price movements. Orabi et al., 2020 reported that there are always low-quality posts, which may skew the performance factor. Thus, investor sentiment should be considered to reduce the influence of poor-quality sentiment. The scientific community has not yet developed a standardized reporting accuracy method. However, the statistical accuracy reported by some authors, as displayed in Table VII.

TABLE VII.  ACCURACY REPORTED BY EARLIER AUTHORS.

| Authors | Accuracy reported | Modelling approach |
|---|---|---|
| Chiang, Enke, Wu, & Wang (2016) | 61.11% | Neural Network |
| Zhong, & Enke, (2017) | 54.48% | Neural Network |
| Sezer & Ozbayoglu, (2018) | 71.51% | Neural Network |
| Hristu-Varsakelis, & Chalvatzis, (2020) | 67.48% | Hybrid Model (Neural Network & Tree Based Algo) |
| Khodaee, Esfahanipour, & Taheri (2022) | 63.62% | Hybrid Neural Network |

Our study outperformed previously reported precision methods, demonstrating that our methodology outperforms existing methodologies in terms of accuracy. This is a noteworthy accomplishment because the goal is to create accurate prediction results and risk-adjusted profits using simple algorithms with low data needs, which is consistent with the suggestions of other researchers, such as Zhong & Enke, 2019.

To this end, our work incorporates an FD series, enabling the modelling of both short- and long-term dependencies in time-series analysis while preserving autocorrelation structures. We leverage sentiment ratings to enhance the accuracy of machine learning models in predicting stock price fluctuations, highlighting the importance of sentiment analysis in financial forecasting. Additionally, we mitigated the impact of low-quality posts by incorporating sentiment data quality assessment, resulting in more reliable outcomes. By advocating the use of Matthews Correlation Coefficient (MCC) scores as a comprehensive evaluation metric for classifier performance, our study adds standardized reporting accuracy approaches. These contributions advance our understanding of efficient methods for classification accuracy and financial forecasting by displaying improved accuracy, highlighting the benefits of FD series, underscoring the importance of sentiment scores, and introducing MCC scores as a robust evaluation metric.

## VI. CONCLUSION

This study demonstrated the effectiveness of fractional differencing (FD) in minimizing long-term dependencies while preserving short-term dependencies in price series. Through this process, the study found that the FD series significantly improves the accuracy of the empirical data models compared with the integer differenced series. Using a difference order of 0.3 for the SPY series, which exhibits stationary properties and a high correlation (>90%) with the original series, the study revealed the presence of long memory. Various supervised classification algorithms, including LogReg, KNN, SVM, and RF are experimented in this work to demonstrate the overall improvements in classification tasks. The accuracy measures have improved noticeably, particularly the ROCAUC (Receiver Operating Characteristic Area Under the Curve) and MCC (Matthews Correlation Coefficient) values. This emphasized the importance of careful time-series modeling and advises against using the default stationarity format (d=1) without considering the statistical properties of the data. Although the current findings are promising, further improvements and advancements can be made in future research. This empirical investigation adds original insights to time-series modeling.

REFERENCES

[1] Alves, P. R. L. (2019). Chaos in historical prices and volatilities with five-dimensional Euclidean spaces. *Chaos, Solitons & Fractals*: X, 1, 100002.

[2] Asl, M. S., & Javidi, M. (2017). An improved PC scheme for nonlinear fractional differential equations: Error and stability analysis. Journal of Computational and Applied Mathematics, 324, 101-117.

[3] Ayadi, O. F., Williams, J., & Hyman, L. M. (2009). Fractional dynamic behavior in Forcados Oil Price Series: An application of detrended fluctuation analysis. Energy for Sustainable Development, 13(1), 11-17.

[4] Baillie, R. T. (1996). Long memory processes and fractional integration in econometrics. Journal of econometrics, 73(1), 5-59.

[5] Beran, J. (1994). Statistics for long-memory processes (Vol. 61). CRC press.

[6] Bozanta, A., Angco, S., Cevik, M., & Basar, A. (2021, December). Sentiment Analysis of StockTwits Using Transformer Models. In 2021 20th *IEEE International Conference on Machine Learning and Applications* (ICMLA) (pp. 1253-1258). IEEE.

[7] Cavaliere, G., Nielsen, M. Ø., & Robert Taylor, A. M. (2022). Adaptive inference in heteroscedastic fractional time series models. *Journal of Business & Economic Statistics*, 40(1), 50-65.

[8] Cavaliere, G., Nielsen, M. Ø., & Taylor, A. R. (2017). Quasi-maximum likelihood estimation and bootstrap inference in fractional time series models with heteroskedasticity of unknown form. *Journal of Econometrics*, 198(1), 165-188.

[9] Cerqueti, R. and Rotundo, G., 2015. A review of aggregation techniques for agent-based models: understanding the presence of long-term memory. Quality & Quantity 49, 1693-1717.

[10] Cerqueti, R., Rotundo, G., 2012. The Role of Diversity in Persistence Aggregation. International Journal of Intelligent Systems 27, 176-187.

[11] Chang, J., Tu, W., Yu, C., & Qin, C. (2021). Assessing dynamic qualities of investor sentiments for stock recommendation. *Information Processing & Management*, 58(2), 102452.

[12] Chen, X., Zhu, H., Zhang, X., & Zhao, L. (2022). A novel time-varying FIGARCH model for improving volatility predictions. Physica A: Statistical Mechanics and its Applications, 589, 126635.

[13] Chiang, W. C., Enke, D., Wu, T., & Wang, R. (2016). An adaptive stock index trading decision support system. *Expert Systems with Applications*, 59, 195-207.

[14] Chicco, D., Warrens, M. J., & Jurman, G. (2021). The Matthews correlation coefficient (MCC) is more informative than Cohen's Kappa and Brier score in binary classification assessment. *IEEE Access*, 9, 78368-78381.

[15] David, S. A., Machado, J. A., Trevisan, L. R., Inacio Jr, C., & Lopes, A. M. (2017). Dynamics of commodities prices: integer and fractional models. Fundamenta Informaticae, 151(1-4), 389-408.

[16] De Prado, M. L. (2018). Advances in financial machine learning. John Wiley & Sons.

[17] Diebold, F. X., & Inoue, A. (2001). Long memory and regime switching. Journal of econometrics, 105(1), 131-159.

[18] Doukhan, P., Oppenheim, G., & Taqqu, M. (Eds.). (2002). Theory and applications of long-range dependence. Springer Science & Business Media.

[19] Flores-Muñoz, F., Báez-García, A. J., & Gutiérrez-Barroso, J. (2018). Fractional differencing in stock market price and online presence of global tourist corporations. *Journal of Economics, Finance and Administrative Science*, 24(48), 194-204.

[20] Gil-Alana, L. A., Yaya, O. S., & Awe, O. O. (2017). Time series analysis of co-movements in the prices of gold and oil: Fractional cointegration approach. *Resources Policy*, 53, 117-124.

[21] Granger, C. W., & Ding, Z. (1995). Some properties of absolute return: An alternative measure of risk. Annales d'Economie et de Statistique, 67-91.

[22] Granger, C. W., & Joyeux, R. (1980). An introduction to long-memory time series models and fractional differencing. *Journal of time series analysis*, 1(1), 15-29.

[23] Haubrich, J. G. (1993). Consumption and fractional differencing: Old and new anomalies. The Review of Economics and Statistics, 767-772.

[24] Hosking, J.R.M., (1981). Fractional differencing. *Biometrika* 68 (1), 165–176.

[25] Hristu-Varsakelis, D., & Chalvatzis, C. (2020). High-performance stock index trading via neural networks and trees.

[26] Hurst, H. E. (1951). Long-term storage capacity of reservoirs. Transactions of the American society of civil engineers, 116(1), 770-799.

[27] Hurst, H. E. (1957). A suggested statistical model of some time series which occur in nature. Nature, 180(4584), 494-494.

[28] Jensen, A. N., & Nielsen, M. Ø. (2014). A fast fractional difference algorithm. *Journal of Time Series Analysis*, 35(5), 428-436.

[29] Johansen, S., & Nielsen, M. Ø. (2010). Likelihood inference for a nonstationary fractional autoregressive model. *Journal of Econometrics*, 158(1), 51-66.

[30] Johansen, S., & Nielsen, M. Ø. (2019). Nonstationary cointegration in the fractionally cointegrated VAR model. *Journal of Time Series Analysis*, 40(4), 519-543.

[31] Kapetanios, G., Papailias, F., & Taylor, A. R. (2019). A generalised fractional differencing bootstrap for long memory processes. *Journal of Time Series Analysis*, 40(4), 467-492.

[32] Khodaee, P., Esfahanipour, A., & Taheri, H. M. (2022). Forecasting turning points in stock price by applying a novel hybrid CNN-LSTM-ResNet model fed by 2D segmented images. *Engineering Applications of Artificial Intelligence*, 116, 105464.

[33] Kitagawa, G. (1987). Non-gaussian state—space modeling of nonstationary time series. Journal of the American statistical association, 82(400), 1032-1041.

[34] Landis, J.R.; Koch, G.G. (1977). "The measurement of observer agreement for categorical data". *Biometrics* 33 (1): 159–174.

[35] Mackinnon, J.G., 1996. Numerical distribution functions for unit root and cointegration tests. *J. Appl. Econ*. 11, 601–618.

[36] Mandelbrot, B. B., & Wallis, J. R. (1968). Noah, Joseph, and operational hydrology. Water resources research, 4(5), 909-918.

[37] McLeod, A. I., & Hipel, K. W. (1978). Preservation of the rescaled adjusted range: 1. A reassessment of the Hurst Phenomenon. Water Resources Research, 14(3), 491-508.

[38] Mikosch, T., & Stărică, C. (2004). Nonstationarities in financial time series, the long-range dependence, and the IGARCH effects. Review of Economics and Statistics, 86(1), 378-390.

[39] Mills, T. C. (2019). Applied time series analysis: A practical guide to modeling and forecasting. Academic press.

[40] Nguyen, H. P., Liu, J., & Zio, E. (2020). A long-term prediction approach based on long short-term memory neural networks with automatic parameter optimization by Tree-structured Parzen Estimator and applied to time-series data of NPP steam generators. Applied Soft Computing, 89, 106116.

[41] Nielsen, M. Ø., & Popiel, M. K. (2014). A Matlab program and user's guide for the fractionally cointegrated VAR model (No. 1330). *Queen's Economics Department Working Paper*.

[42] Obaid, K., & Pukthuanthong, K. (2022). A picture is worth a thousand words: Measuring investor sentiment by combining machine learning and photos from news. *Journal of Financial Economics*, 144(1), 273-297.

[43] Orabi, M., Mouheb, D., Al Aghbari, Z., & Kamel, I. (2020). Detection of bots in social media: a systematic review. *Information Processing & Management*, 57(4), 102250.

[44] Pan, Q., Li, P., & Du, X. (2023). An improved FIGARCH model with the fractional differencing operator (1-vL) d. Finance Research Letters, 103975.

[45] Raudys, A., & Goldstein, E. (2022). Forecasting Detrended Volatility Risk and Financial Price Series Using LSTM Neural Networks and XGBoost Regressor. Journal of Risk and Financial Management, 15(12), 602.

[46] Robinson, P. M. (1995). Log-periodogram regression of time series with long range dependence. The annals of Statistics, 1048-1072.

[47] Sadaei, H. J., Enayatifar, R., Guimarães, F. G., Mahmud, M., & Alzamil, Z. A. (2016). Combining ARFIMA models and fuzzy time series for the forecast of long memory time series. Neurocomputing, 175, 782-796.

[48] Serinaldi, F. (2010). Use and misuse of some Hurst parameter estimators applied to stationary and non-stationary financial time series. Physica A: Statistical Mechanics and its Applications, 389(14), 2770-2781.

[49] Sezer, O. B., & Ozbayoglu, A. M. (2018). Algorithmic financial trading with deep convolutional neural networks: Time series to image conversion approach. *Applied Soft Computing*, 70, 525-538.

[50] Smith, W., & Harris, C. M. (1987). Fractionally differenced models for water quality time series. Annals of Operations Research, 9, 399-420.

[51] Sun, L., Najand, M., & Shen, J. (2016). Stock return predictability and investor sentiment: A high-frequency perspective. *Journal of Banking & Finance*, 73, 147-164.

[52] Wang, Q., & Xu, R. (2022). A review of definitions of fractional differences and sums. *Mathematical Foundations of Computing*, 0-0.

[53] Yang, S. Y., Yu, Y., & Almahdi, S. (2018). An investor sentiment reward-based trading system using Gaussian inverse reinforcement learning algorithm. *Expert Systems with Applications*, 114, 388-401.

[54] Zaffaroni, P., 2007. Memory and aggregation for models of changing volatility. Journal of Econometrics 136, 237-249.

[55] Zhang, Y., Song, W., Karimi, M., Chi, C. H., & Kudreyko, A. (2018). Fractional autoregressive integrated moving average and finite-element modal: the forecast of tire vibration trend. IEEE Access, 6, 40137-40142.

[56] Zhong, X., & Enke, D. (2017). Forecasting daily stock market return using dimensionality reduction. *Expert Systems with Applications*, 67, 126-139.

[57] Zhong, X., & Enke, D. (2019). Predicting the daily return direction of the stock market using hybrid machine learning algorithms. *Financial Innovation*, 5(1), 1-20.